\documentclass[conference]{IEEEtran}
\IEEEoverridecommandlockouts
\usepackage{cite}
\usepackage{amsmath,amssymb,amsfonts}
\usepackage{algorithmic}
\usepackage{comment}
\usepackage{graphicx}
\usepackage{textcomp}
\usepackage{xcolor}
\usepackage{multirow}
\usepackage{enumitem}
\usepackage{subfig}
\usepackage{hyperref}
\usepackage{balance}

\def\BibTeX{{\rm B\kern-.05em{\sc i\kern-.025em b}\kern-.08em
    T\kern-.1667em\lower.7ex\hbox{E}\kern-.125emX}}
\begin{document}

\title{REFLEX Dataset: A Multimodal Dataset of Human Reactions to Robot Failures and Explanations
\\
}
%REFLEX Dataset: Robtic explanations to failues and Human expressions dataset

\author{
\hspace{-60pt}\IEEEauthorblockN{Parag Khanna}
\IEEEauthorblockA{%\textit{Division of Robotics, Perception and Learning (RPL)} \\
\hspace{-60pt}\textit{KTH Royal Institute of Technology}\\
\hspace{-60pt}Stockholm, Sweden \\
\hspace{-60pt}paragk@kth.se}
\and
\IEEEauthorblockN{Andreas Naoum}
\IEEEauthorblockA{%\textit{Division of Robotics, Perception and Learning (RPL)} \\
\textit{KTH Royal Institute of Technology}\\
Stockholm, Sweden\\
anaoum@kth.se}
\and
\IEEEauthorblockN{Elmira Yadollahi}
\IEEEauthorblockA{\textit{Lancaster University} \\
Lancaster, United Kingdom \\
e.yadollahi@lancaster.ac.uk}
\and \hspace{100pt}\IEEEauthorblockN{Mårten Björkman}
\IEEEauthorblockA{%\textit{Robotics, Perception and Learning (RPL)} \\
\hspace{100pt}\textit{KTH Royal Institute of Technology}\\
\hspace{100pt}Stockholm, Sweden\\
\hspace{100pt}celle@kth.se}
\and
\IEEEauthorblockN{Christian Smith}
\IEEEauthorblockA{%\textit{Robotics, Perception and Learning (RPL)} \\
\textit{KTH Royal Institute of Technology}\\
Stockholm, Sweden \\
ccs@kth.se}
}
\maketitle
\begin{abstract}
This work presents REFLEX: Robotic Explanations to FaiLures and Human EXpressions, a comprehensive multimodal dataset capturing human reactions to robot failures and subsequent explanations in collaborative settings. 
It aims to facilitate research into human-robot interaction dynamics, addressing the need to study reactions to both initial failures and explanations, as well as the evolution of these reactions in long-term interactions. By providing rich, annotated data on human responses to different types of failures, explanation levels, and explanation varying strategies, the dataset contributes to the development of more robust, adaptive, and satisfying robotic systems capable of maintaining positive relationships with human collaborators, even during challenges like repeated failures.
\end{abstract}


\section{Introduction}
As robots become increasingly integrated into our everyday lives, from homes and workplaces to public spaces, the need to understand and improve human-robot interaction (HRI) has never been more critical. Despite significant advancements in robotics, they are still prone to failures, ranging from minor glitches to serious malfunctions. When robots fail, particularly while collaborating with humans, it's critical that they provide apt explanations for failure to their human collaborators, allowing for quick resolution and sustaining human trust.

Studying human reactions to robotic failures is crucial for several reasons. First, it helps in developing more effective HRI systems by anticipating and addressing potential issues \cite{tabrez2019reactive}. Second, understanding these reactions allows the creation of tailored explanations that address specific user concerns \cite{kwon2018expressing}, helping to maintain appropriate trust levels \cite{wang2016impact}. Third, it improves collaboration by enabling robots to anticipate and respond to human reactions more effectively \cite{tabrez2019reactive}. Lastly, it enhances the overall user experience by considering both emotional and cognitive responses to failures \cite{spitale2024err}.
Moreover, the study of human reactions to robotic explanations of failures is equally important. While explanations have been shown to improve transparency and trust calibration \cite{wang2016impact}, their effectiveness can vary based on the specific failure context and the individual user \cite{exp_strategies_roman_khanna2023}. Understanding how humans respond to different types of explanations can lead to more nuanced and effective explanation strategies for robots.

Further, there is a need to investigate human reactions to repeated failures and explanations. As robots become more prevalent in long-term interactions \cite{desai2013impact,salem2015err}, it's crucial to understand how trust and collaboration dynamics evolve over time, especially in the face of recurring issues. This understanding can inform the development of adaptive explanation strategies that maintain positive human-robot relationships even in challenging circumstances.
To address these research requirements, this work presents a comprehensive multimodal dataset capturing human reactions to both robot failures and the explanations provided for these failures. This dataset \cite{dataset_zenodo} goes beyond existing resources by including reactions to various levels and strategies of explanations, as well as responses to repeated failures and explanations over time. By providing this rich, annotated data, we aim to facilitate further research into HRI dynamics, ultimately contributing to the development of more adaptive and effective HRI systems.
%more robust, trustworthy, and satisfying robotic systems.
\begin{figure}[t]
      \centering
        \includegraphics[width=.45\linewidth,height=2.8cm,trim={0.0cm 0.0cm 0.0cm 0.0cm},clip]{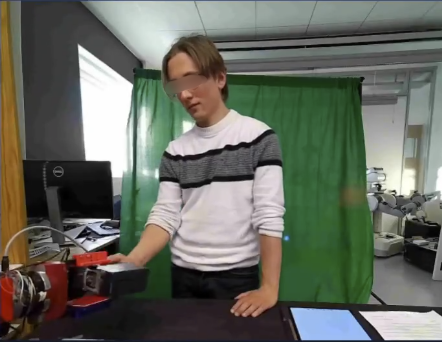} 
        \includegraphics[width=.45\linewidth,height=2.8cm,trim={0.0cm 0.0cm 0.0cm 0.0cm},clip]{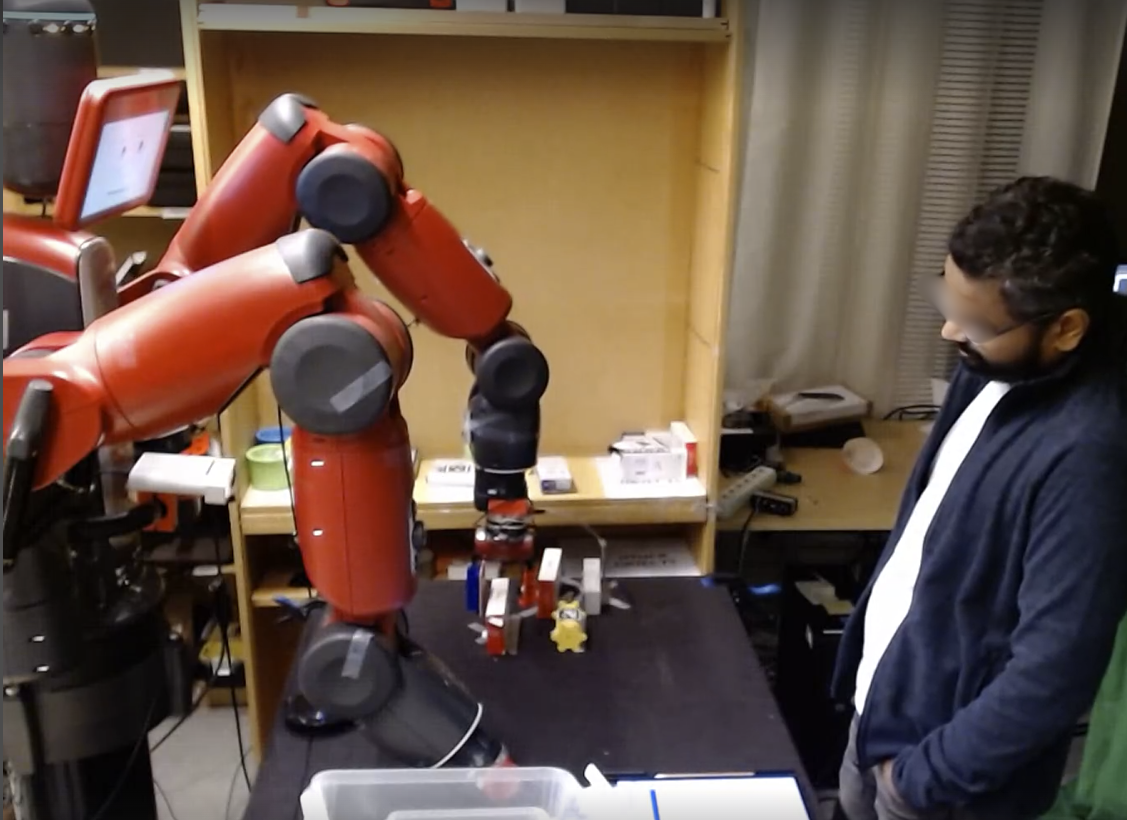}  
     %trim={left bottom right top}
    %\setlength\abovecaptionskip{-0.05\baselineskip}
    \caption{Human Robot Collaboration (HRC) task captured by 2 cameras: Camera 1 (left) focused on the user \& Camera 2.}
    %\description[2 camera view]{Figure depicts the human-robot collaboration task in our experiment being recorded by two cameras. The left image displays the output from Camera 1, which focuses on the user. The right image shows the output from Camera 2, capturing both the user and the Baxter robot, thus encompassing the entire interaction.}
    \label{fig:HRC_task_view}
\end{figure}
 %``Modelling human reaction"
\section{Related work}
Research on human reactions to robotic failures and robotic explanations of failures has gained increasing attention in recent years. 
Understanding human reactions to robot failures is crucial for developing effective HRI systems. Several studies have examined how humans perceive and respond to robot errors in collaborative tasks. \cite{mirnig2017understanding} found that humans tend to attribute robot failures to technical issues rather than the robot's cognitive capabilities. However, \cite{desai2013impact} showed that repeated failures can negatively impact trust and acceptance of robots. The type and severity of failures also influence human perceptions, with task-related errors being more detrimental than social norm violations \cite{salem2015err}. Further, providing explanations for failures has been shown to improve human-robot interaction. \cite{wang2016impact} demonstrated that explanations can increase transparency and help calibrate trust in robotic systems. \cite{thielstrom2020generating} proposed an approach for generating explanations about action failures in cognitive robotic architectures. Recent work by \cite{liu2023reflect} has explored using large language models to generate informative failure explanations for robots.
Studying human reactions to robotic failures is also crucial for the creation of tailored explanations that address specific user concerns \cite{kwon2018expressing}, help maintain appropriate trust levels \cite{wang2016impact}, improve collaboration by anticipating and addressing reactions \cite{tabrez2019reactive}, and enhance the overall user experience by considering emotional and cognitive responses \cite{spitale2024err}. By incorporating these insights, explanation strategies can be designed, leading to more robust, trustworthy, and satisfying HRI, even when failures occur.

However, only a few publicly available datasets exist documenting human reactions to robotic failures, specifically those involving real robot-human interactions rather than humans viewing robot videos.
\cite{stiber2023react} introduced the REACT dataset, containing multimodal data of human reactions to various types of robot failures in a collaborative task.
Another dataset, not yet publicly available, ERR@HRI 2024 challenge dataset \cite{spitale2024err} provides multimodal non-verbal interaction data, including facial, speech, and pose features from interactions with a robotic coach, annotated with labels of robot mistakes and user reactions. However, these datasets primarily address the initial failure reactions, neglecting the human reactions after the robot explains the failure. As discussed in \cite{tabrez2019reactive,spitale2024err}, the explanations can be tailored not only based on specific human reactions to failures but also the human reactions to previous explanations provided by the robot for similar failure. Our dataset not only thoroughly documents the diverse human reactions to various robot failures but also captures the responses to different levels and strategies of explanations provided by the robot.

\section{Data Collection}
Audio-visual data was collected as part of a user study in a prior work \cite{khanna2023userstudyexploringrole,exp_strategies_roman_khanna2023} where users collaborated with a robot. The setup for the study is as described in Fig. \ref{fig:HRC_task_view}. In the HRC task, the users kept objects on a table in front of the robot, and the robot would keep the objects on a shelf. The users interacted with the robot in four rounds of four objects each. The robot needs to do a series of actions to handle each object: Detect-$>$Pick-$>$Carry-$>$Place the object on shelf. To induce robotic failures for certain objects, robotic action failures were pre-programmed for the Pick, Carry and Place actions. 4 instances for each failure type occurred across the 4 rounds, for 9 out of 16 objects. At each failure, the robot would provide an explanation for the failure with the required resolution from the user. These failures repeated in different rounds of interactions, and the explanation level for the robot was according to the set explanation strategy, shown in Table 1. We set 5 strategies, with \textit{Fixed} strategies keeping a fixed level of explanation in the four rounds, while \textit{Decay} strategies had a reduced level of explanation in subsequent rounds, shown in Table 1. 
\begin{table}[t]
    \centering
    \setlength{\abovecaptionskip}{-0.125\baselineskip}
    \caption{Explanation Strategies}
    \label{tab:round-results}
    \scalebox{0.895}{
    \begin{tabular}{|c|l|c|c|c|c|}
        \hline
        ID & Details & Round 1 & Round 2 & Round 3 & Round 4 \\
        \hline
        C1 & Fixed-Low & Low & Low & Low & Low \\
        C2 & Fixed-Medium & Mid & Mid & Mid & Mid \\
        C3 & Fixed-High & High & High & High & High \\
        D1 & Decay-Slow & High & Mid & Low & None \\
        D2 & Decay-Rapid & High & Low & Low & Low \\
        \hline
    \end{tabular}
    }
    \vspace{-1mm}
\end{table}
The explanation levels considered are as follows:
\begin{itemize}[leftmargin=*]
    \item Zero Level, Non-verbal explanation: The robot shakes its head and goes into a handover pose after each failure.
    \item Low, Action-based: The robot states the failure and asks for help.
   ``I failed to pick up the object", ``Hand it to me".
   %- "I failed to carry the object", "Carry it for me".\\
   %- "I failed to place the object", "Place it for me".
   \item Medium, Context-based: The robot explains the failure cause and asks for help.
   ``I can't pick up the object because it doesn't fit in my gripper", ``Can you hand it over to me".
   %- "I can't carry the object because it is too heavy for my arm", "Can you carry it for me".\\
   %- "I can't place the object because the destination is out of my arm's reach", "Can you place it for me".
   \item High, Context + History-based: The robot mentions a previous success, explains the current failure and its cause, and asks for help.
   ``I can detect the object, but I can't pick it up because it doesn't fit in my gripper", ``Can you hand it over to me by placing it in my gripper?".
   %- "I can pick up the object, but I can't carry it because it is too heavy for my arm", "Can you carry the object for me and place it on the shelf?".\\
   %- "I can carry the object, but I can't place it because the destination is out of my arm's reach", "Can you place the object on the shelf location that is out of my reach?".
\end{itemize}
The raw collected data included the audio recording of the interaction and the video recordings from 2 cameras in Fig. \ref{fig:HRC_task_view}: a camera focused on both the user and the robot to cover the interaction; and a camera placed on the torso of the robot focused solely on the user. After each round, the users answered 8 questions assessing explanation satisfaction based on \cite{hoffman2018metrics}, evaluating understandability, satisfaction, detail sufficiency, completeness, usefulness, accuracy, and trustworthiness. 
\begin{table*}[t]
    \centering
       \setlength{\abovecaptionskip}{-0.125\baselineskip}
    \caption{Dataset Components}
    \renewcommand{\arraystretch}{1.10} % Adjust the value to increase/decrease the space
    \scalebox{0.9}{
    \begin{tabular}{|p{0.01cm} p{0.05cm}|p{1.8cm}|p{3.8cm}|p{8.7cm}|p{2.0cm}|}
        \hline
        %\multicolumn{4}{|l|}{\textbf{Experiment Data}} \\ %\hline
        \multirow{5}{*}{\rotatebox[origin=c]{90}{\textbf{Experiment}}} & \multirow{5}{*}{\rotatebox[origin=c]{90}{\textbf{Data}}} & \textbf{Component} & \multicolumn{3}{p{14.2cm}|}{\textbf{Description}}  \\ \cline{3-6}
        & &
        Visual Representation & \multicolumn{3}{p{14.2cm}|}{Human-robot interaction is visually represented by segmenting the user with a black mask to ensure anonymity from Camera 1, Camera 2 view} \\ \cline{3-6}
        & & Failure Instance Description & \multicolumn{3}{p{13.2cm}|}{ Failure Action (Pick, Place or Carry), Explanation level (High, Medium or Low), task resolution outcome (Success or Failure), Explanation Strategy, Round 1, Phase (Pre, Failure, Explanation, Resolution), Start/End Frame for the Phase, Start/End Time for the Phase, Explanation-Satisfaction Responses from the participant} \\ \hline
        %\multicolumn{3}{|l|}{\textbf{Participant Data}} \\ \hline
        \multirow{15}{*}{\rotatebox[origin=c]{90}{\textbf{Participant}}} & \multirow{15}{*}{\rotatebox[origin=c]{90}{\textbf{Data}}} & \textbf{Modality} & \textbf{Component} & \textbf{Description} & \textbf{Tool} \\ \cline{3-6}
        & & \multirow{2}{*}{Speech} 
        & Verbal Exchange Transcription & Complete transcription of verbal exchange in experiment & Hume \\ \cline{4-6}
        & & & Speech Prosody based Emotions & Emotion likelihood based on speaker's voice & Hume \\ \cline{3-6}
        & & \multirow{5}{*}{Face}
        & Facial Landmarks & Facial Landmarks in 2D and 3D as normalized values & OpenFace \\ \cline{4-6}
        & & & Facial Action Units & Occurrences and Intensities of Facial Muscle Movements & OpenFace, Hume \\ \cline{4-6}
        & & & Facial Descriptions & Intensities of Facial Descriptions (e.g. Smile) & Hume \\ \cline{4-6}
        & & &Facial Emotions & Likelihood values of 48 distinct emotions & Hume \\ \cline{4-6}
        & & & Affective State & Arousal/Valence scores, Strongest Emotion & Facetorch \\ \cline{3-6}
        & & Gaze
        & Eye Gaze Landmarks & Eye Gaze Landmarks in 2D and 3D  & OpenFace \\ \cline{4-6}
        & & (Normalized) & Eye Gaze Direction & Eye gaze direction vector (x,y,z) and  direction in radians (x,y) & OpenFace \\ \cline{4-6}
        & & & Gaze Classification & Gaze on Robot, Task, Misc. & Annotated \\ \cline{3-6}
        & & \multirow{2}{*}{Head} 
        & Pose Estimation & Location of the head with respect to camera (x,y,z) & OpenFace \\ \cline{4-6}
        & & & Rotation & Rotation is in radians around X,Y,Z axes (pitch, yaw, roll) & OpenFace \\ \cline{3-6}
        & & \multirow{2}{*}{Body} 
        & Pose Landmarks & Body Landmarks in 2D and 3D as normalized and world coordinate values & MediaPipe \\ \cline{4-6}
        & & & Pose Classification & Crossed Arms, Arms Behind Back & Annotated \\ \hline
    \end{tabular}
    }
    \vspace{-2mm}
    \label{tab:dataset_components}
\end{table*}

\subsection{Participants}
The participants for the study were recruited via advertisement on the campus. As a necessary prerequisite, we selected participants who had no prior experience in physically interacting with a robot. 
We selected 11 participants per strategy: N = 55 (age M=26.63, SD=7.42), 21 Female, 33 Male, 1 Other.
%\( N = 55 \) (\( M = 26.63, SD = 7.42 \)) (21 Female, 33 Male, 1 Other). 
As per the local regulations, we are exempt from ethical approval as we did not collect any sensitive personal data (racial/ethnic origin, political views, religious/philosophical beliefs, health/sexual life) and this research doesn't involve physical intervention on or biological samples from participants.
In the absence of a relevant ethics board, we followed guidelines of the Declaration of Helsinki.
Participants began by completing a consent form for data collection; and reading the study instructions. 
Particularly, they consented to the use and distribution of their anonymized data and the use of collected video data %during the experiment 
in academic articles and presentations. 
They were informed about their role in placing objects on the table and the robot's role in picking and placing objects on the shelf; but, potential failures and resolutions were not mentioned. After the experiment, participants received a debriefing sheet explaining the study's aim.  
\begin{figure}[b]
      \centering
        \includegraphics[width=.40\linewidth,height=2.1cm,trim={0.0cm 0.0cm 0.2cm 0.0cm},clip]{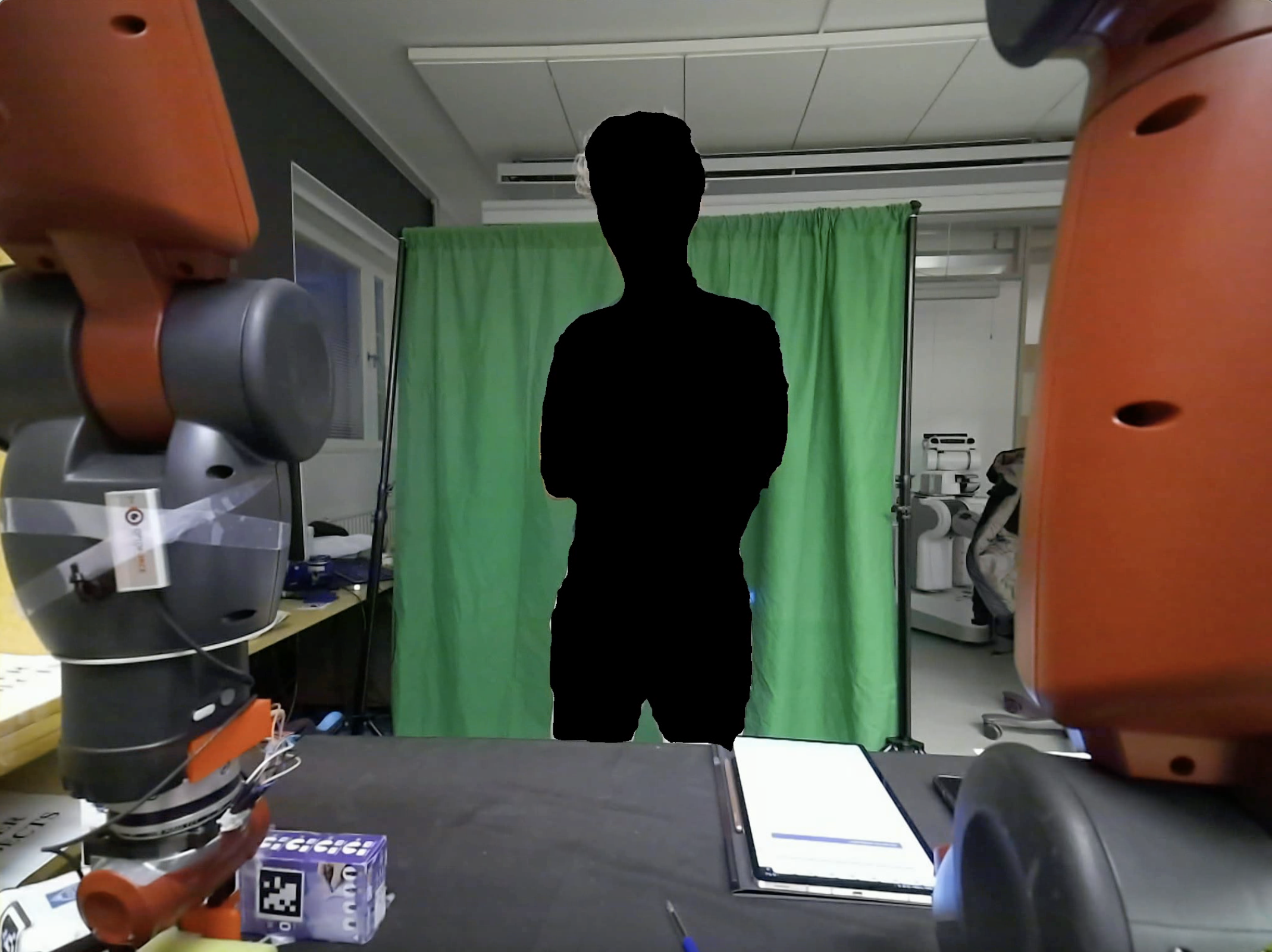} 
        \includegraphics[width=.40\linewidth,height=2.1cm,trim={0.0cm 0.0cm 0.2cm 0.0cm},clip]{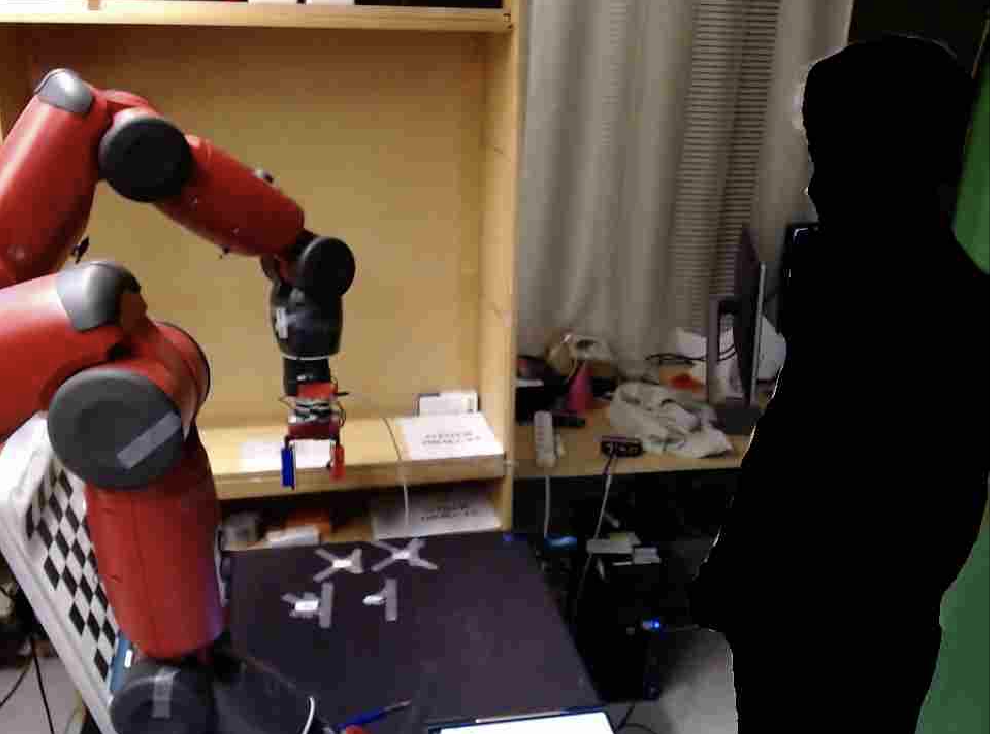}  
     %trim={left bottom right top}
    %\setlength\abovecaptionskip{-0.00\baselineskip}
    \caption{Visual Representation of the HRC Task, user-maskedd view: Camera 1 (left) focused on the user and Camera 2 (right).}
    % \vspace{-1.5mm}
    %\Description{this figure presents the visual representation in the dataset shows the user masked out with a black mask. This approach maintains participant anonymity while still providing visual information about the interaction. The left sub-figure shows the view from Camera 1 with the user replaced by a black mask. The right sub-figure displays the view from Camera 2, also with the user replaced by a black mask.}
    \label{fig:HRC_task_view_cameras}
\end{figure}
\section{Dataset}
%The focus of this dataset is on the human reaction to robotic failures and the explanations provided by the robot. 
This dataset \cite{dataset_zenodo} involves data from 55 participants, 11 each in the 5 explanation strategies in Table 1. 
We process the raw audio-visual data, sampled at a frequency of 4.4 Hz based on the video frame rate, to collect data for each frame of user reactions. In the following, the visual representation is saved in .mp4 format, while all other data is saved in .csv files.

\subsubsection{Visual Representation}
We provide a visual representation of the interaction by segmenting the user out with a black mask, thereby ensuring that the experiment remains visible. This was accomplished using the segmentation mask detected by the MediaPipe Pose Landmark detector \cite{mediapipe_pose}. In cases where no pose was detected and thus no mask was exist, the video displays a gray image to ensure the participant's anonymity. We believe this innovative approach provides valuable insights for HRI while maintaining user anonymization (Fig. \ref{fig:HRC_task_view_cameras}).
\subsubsection{Failure Instance Description}
\begin{figure}[b]
      \centering
      \setlength\abovecaptionskip{-0.05\baselineskip}
      \includegraphics[scale=0.18,trim={2.0cm 15.0cm 1.0cm 0.0cm},clip]{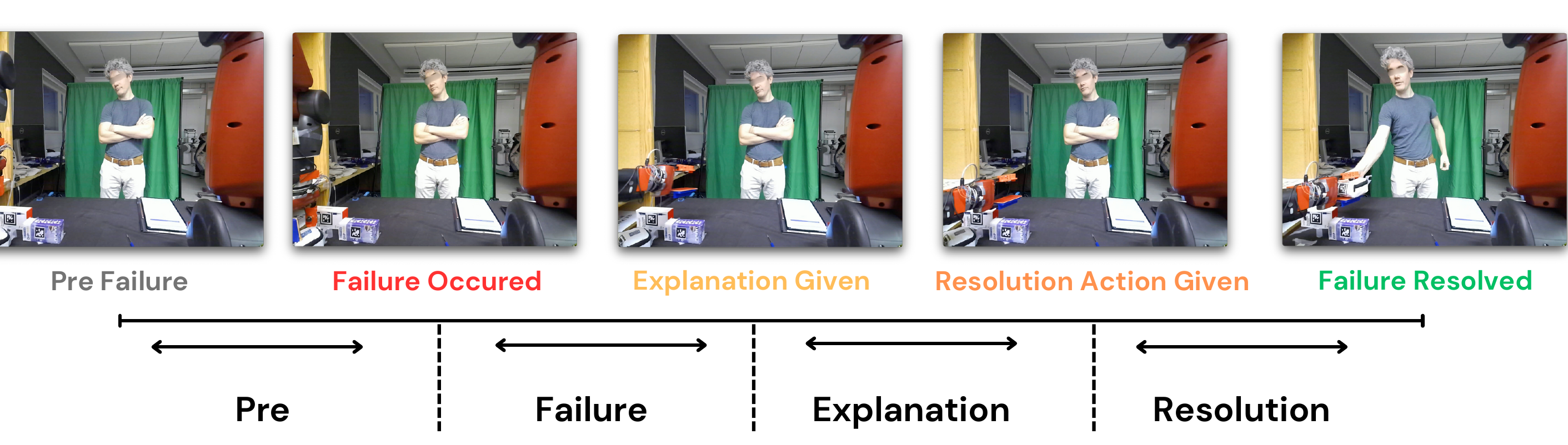}
      \caption{Human Reaction Modeling for Failure Explanation}
      %\Description{This figure is a collection of 5 pictures in temporal occurance depicting the timeline of human reaction to failure, explanation and resoultion with arrows showing the subsequent division into different phases. The first picture shows the pre-failure human reaction, the second picture shows the reaction when the robotic failure occurs, the third picture shows the reaction when explanation is given by the robot, the fourth picture shows the reaction at resolution action given by the robot and the last picture shows the failure resolution by the human. The phases marked by 2 sided arrows are: Pre between first and second, Failure between second and third, Explanation between third and fourth, and Resolution between fourth and fifth.}
      \label{figurelabel3}
      \vspace{-0.5mm}
\end{figure}

Following \cite{What_If_It_Is_Wrong_Karli2023, Model_human_response_to_robot_Error_Stiber2022}, the human reactions are divided into four key phases during a failure event, as shown in Fig. \ref{figurelabel3}. First, there’s the pre-failure phase, which is the period before the failure occurs. Next is the failure phase, where the actual failure action takes place. Then, the explanation phase, during which the robot provides an explanation for the failure. Finally, there’s the resolution phase, where the robot guides the participant to take steps to resolve the issue. Information for each phase of each failure was automatically generated using recorded logs,
%the logs saved in the rosbag, 
which include details such as the failure action, the explanation strategy, the current explanation level, and the start and end frames (times).
\begin{figure*}[thpb]
    \centering
    \subfloat[]{\includegraphics[width=0.24\textwidth]{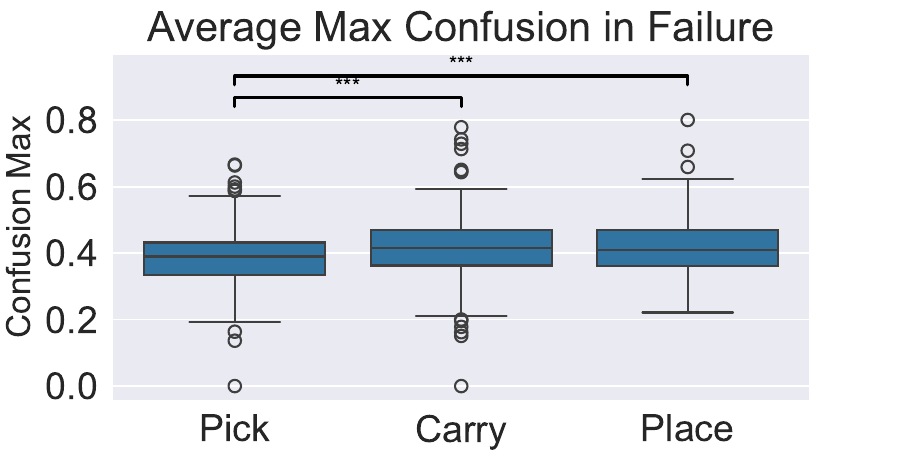}}
    \hfill
    \subfloat[]{\includegraphics[width=0.24\textwidth]{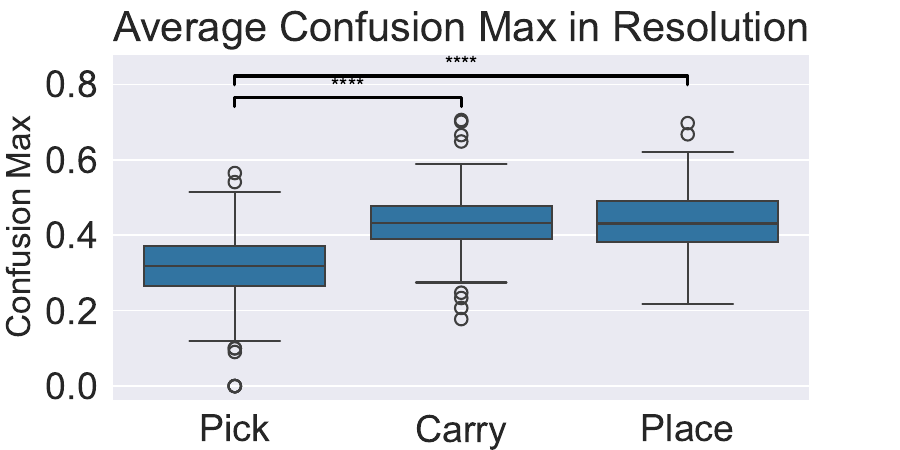}}
    \hfill
    \subfloat[]{\includegraphics[width=0.49\textwidth,height=2.05cm]{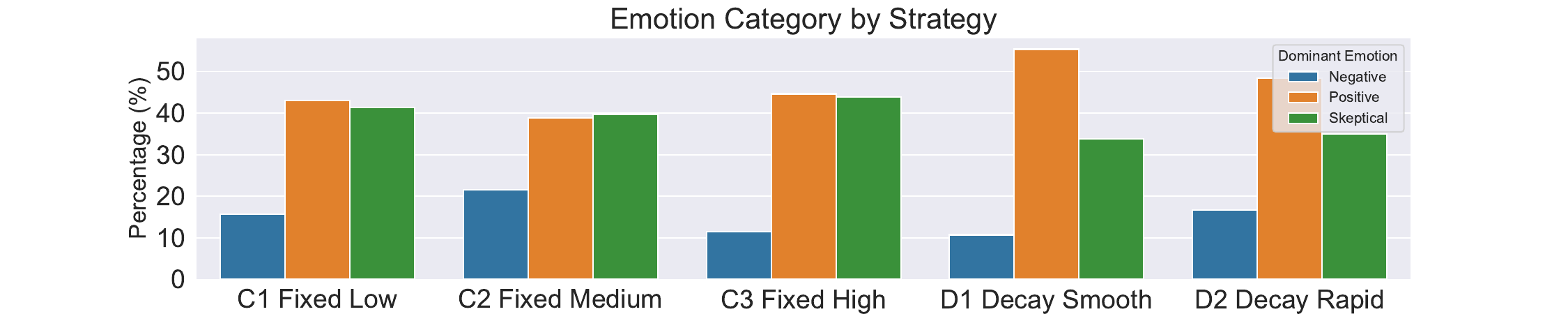}}
    \setlength\abovecaptionskip{-0.05\baselineskip}
    \caption{ Analyzing Human Reactions by Detected Facial Emotions. Reaction to Different Robotic Failures by comparing likelihood of confusion in: (a) Failure Phase and (b) Resolution Phase, (c) Reaction to Different Robotic Explanations for Failures, Dominant emotional category seen across different explanation strategies in explanation and resolution phase}
    \label{figure4}
    \vspace{-1mm}
\end{figure*}
\subsubsection{Speech}
During the experiment, participants verbally interacted with the robot.
%received statements, questions and guides. 
By using the Hume Expression Measurement API for Speech Prosody \cite{hume_api,hume_speech_prosody}, the conversation was transcribed into sentences attributed to the identified speaker. Each sentence was analyzed for the likelihood ($\mathcal{L}$) of 48 distinct emotions expressed through the speaker's voice, considering non-linguistic elements such as tone, rhythm, and timbre of speech. The $\mathcal{L}$ metric provides a quantitative measure of emotional intensity and expressiveness.
% All the conversation is transcribed and the speaker is Speech Prosody encompasses the non-linguistic tone, rhythm, and timbre of speech. These can be viewed as emotions expressed by the speaker voice, based on processing from HumeAI API, spanning 48 distinct dimensions of emotional meaning.
\subsubsection{Face}
There is a widespread agreement that ``the face is a rich source of information”, which motivated us to gather extensive data. Initially, we obtained fundamental yet essential facial data using OpenFace \cite{openface}, including 2D and 3D facial landmarks \cite{openface_landmarks} and the intensity and presence of Facial Action Units (FAUs) \cite{openface_faus}. Recognizing the significance of emotion recognition in understanding participants' affective states during human-robot interaction \cite{cowie2001emotion}, we utilized Facetorch \cite{facetorch} to detect affected states such as arousal and valence \cite{facetorch_arousal}, and the strongest emotion out of the six basic emotions (including Neutral) \cite{facetorch_fer} supported by the presence of FAUs. In light of recent research indicating that facial expressions are complex and high-dimensional, conveying at least 28 dimensions of emotional meaning, we leveraged the Hume Expression Measurement API \cite{hume_api, hume_face} to gather additional data, including $\mathcal{L}$ values for 48 emotions based on facial expressions, as well as detected scores for FAUs and facial descriptions such as ``Smile," ``Hand Touching Face/Head," ``Frown" and others. 
%It is worth mentioning that the FAUs from Hume are based on MediaPipe's Face Landmark Detection.
All 
%these datasets 
the facial data include raw output values for each frame. 

\subsubsection{Gaze}
Gaze plays a critical role in focus, attention, and engagement during HRI. We extracted eye gaze data using OpenFace \cite{openface, openface_eyes} for each frame, including landmarks and gaze directions. For the failure phase, we automatically annotated (followed by manual validation) whether the participant’s gaze was directed toward the robot, the task, or other areas.
\subsubsection{Head}
We extracted head-related data using OpenFace\cite{openface} for each frame, which includes the location and rotation of the head. This information is valuable for understanding how head movements may relate to failure and explanation reactions, and gaining insights into non-verbal cues that contribute to the overall understanding of HRI.
\subsubsection{Body}
The MediaPipe Pose Landmark detector \cite{mediapipe_pose} was utilized to detect and extract body pose landmarks along with their visibility and presence in both 2D and 3D for each frame. Due to the structure of our experiment, which only captured the upper body, the obtained body landmark locations were limited to 24 points out of a possible 32. Further, we developed heuristics to identify two specific body poses observed during experiment: crossed arms, 
%determined by the proximity of the hands to the opposite elbow,
and arms positioned behind back.
%indicated when both arms are not visible.
\section{Comprehensive Visualization of the Dataset}
\begin{figure}[b]
      \centering
      \setlength\abovecaptionskip{-0.02\baselineskip}
      \includegraphics[width=.70\linewidth,height=2.8cm,trim={5.0cm 5.0cm 5.0cm 0.0cm},clip]{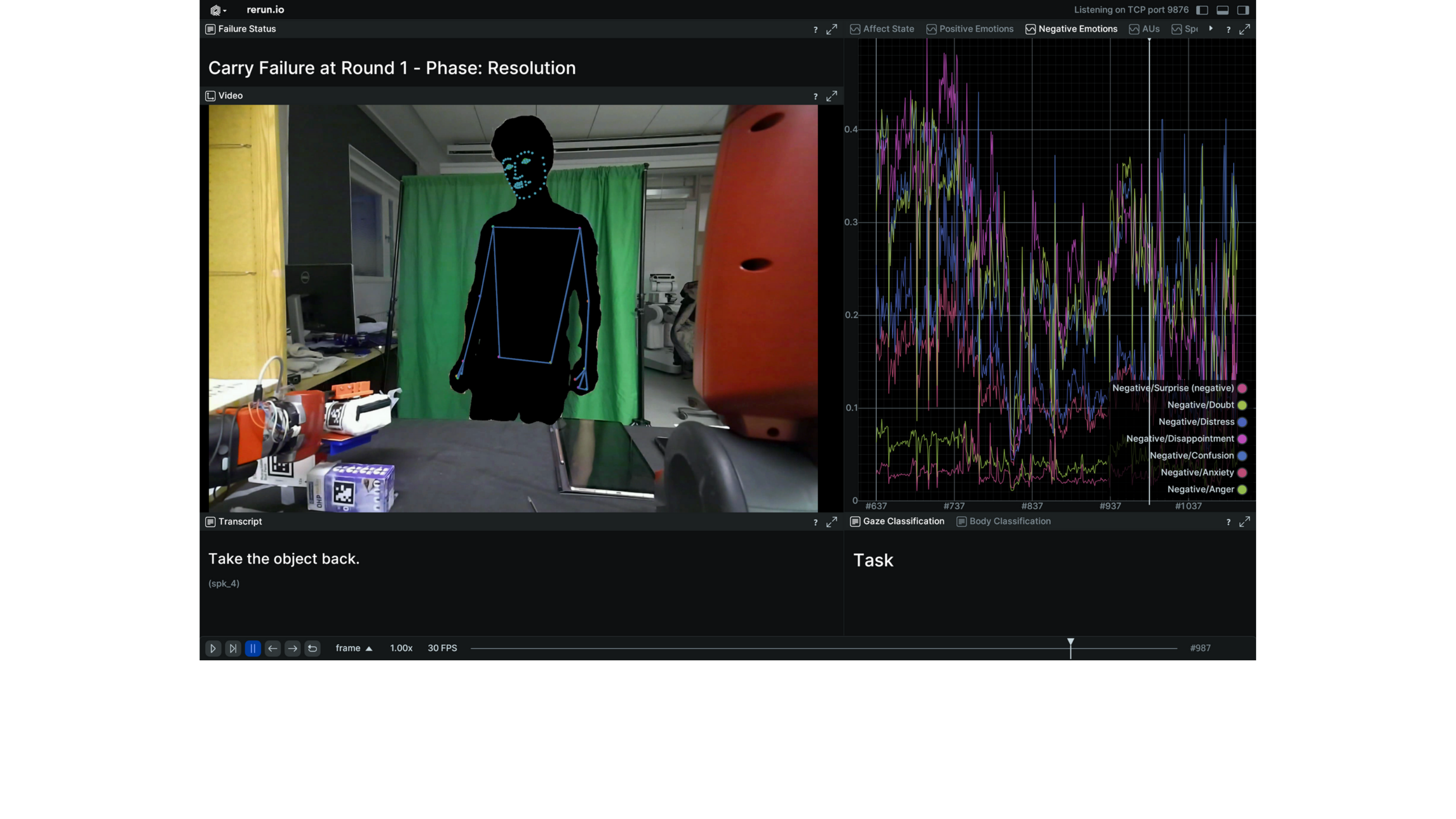}
      \caption{Multimodal Visualization of the Participant's Data}
      \label{figurelabel5}
      %\description{Figure 5: This figure presents a snapshot of the comprehensive multimodal visualization of a participant's data, integrating various data streams with synchronized video footage. The figure has distinct left and right sides. The left side of the visualization displays the anonymized video from camera 1, showing the selected participant. Overlaid on this video are landmarks for the face, body, and eyes, providing a clear visual representation of the participant's physical interactions. Below the video, textual information is presented, including the current failure phase, conversation details, gaze classification, and pose classification. This combination of visual and textual data allows for a comprehensive understanding of the participant's state and actions at any given moment. The right side of the visualization contains graphs displaying various extracted values over time, including emotions, Facial Action Units (FAUs) intensities, and arousal scores. These graphs are temporally aligned with the video footage, enabling viewers to correlate changes in emotional states, facial expressions, and arousal levels with specific events or interactions visible in the video.}
      \vspace{-2.5mm}
\end{figure}

We also provide an easily installable and ready-to-use code-set \cite{dataset_zenodo} for visualizing the interaction with a specific participant, based on Rerun \cite{RerunSDK} open source visualization tool for multimodal data. 
%Users simply need to select the participant to begin the visualization process. 
%A snippet of this visualization is as shown in Fig 4. 
The visualization, Fig. \ref{figurelabel5}, integrates multimodal data with the corresponding video (camera 1) of the selected participant, synchronized by time (or frame), making it easier to understand and interact with the data. Landmarks (face, body, eyes) are overlaid on the video, offering a clear perception of the ongoing interactions. Also, the current failure phase, conversation, gaze, and pose classification are presented as text, while other extracted values such as emotions, FAUs intensities, and arousal scores are displayed in graphs. 

\section{Data Highlights}
In this section, we present just a few selected insights from the dataset. Although the dataset is extensive, we prioritized specific aspects that shed light on participants’ emotional reactions to failures and explanations. 
%  in relation to strategies and actions
We classified certain emotions into three categories: Positive, Negative and Skeptical. The positive category includes the interest, satisfaction, contentment, and desire emotions. Distress, anxiety, anger, and negative surprise fall within the negative category, while confusion and doubt are categorized as skeptical. By calculating both the average and maximum $\mathcal{L}$ of these emotions during the explanation and resolution phases, we were able to determine the dominant category for each instance. We have presented this information in the form of percentages for each category in Fig. \ref{figure4}(c).
Among all the emotions, we consider \textit{confusion} particularly relevant to our experiment, as it offers valuable insights into the complexity of the different failures \cite{identify_int_confusion_LI2023}. Comparing average $\mathcal{L}$ of maximum confusion in failure and resolution phases, Fig. \ref{figure4}(a),(b), we observe a significant difference between pick failure, and carry and place failures, with pick showing lower $\mathcal{L}$ values. This suggests that pick failure was perceived as easier to resolve and less cognitively demanding compared to more complex carry-place failures.

\section{Conclusion}
We present a multimodal dataset of human reactions to different robotic failures, different levels of robotic explanation for these failures and the varying explanation levels in case of repeated failures. We believe that this comprehensive annotated dataset can provide critical insights into designing more robust and adaptable human-robot interaction systems.
It enables the analysis of human responses to various robot failures and explanations, allowing researchers to identify effective approaches for maintaining trust and collaboration. Also, the dataset could be used to develop machine learning models for automatically detecting and classifying human reactions, facilitating more tailored and timely responses from robots. This can also be used to investigate how human reactions evolve over repeated interactions, informing the design of long-term adaptive behaviors.
It is especially beneficial for studying the evaluation of different explanation generation techniques in terms of their impact on human understanding and trust and developing tailored explanations for robotic failures. 
Overall, this dataset is a useful resource for developing socially intelligent and failure-adaptive robotic systems.

\section{Acknowledgment}
%\vspace{-0.5mm}
This work was partially funded by Digital Futures at KTH.
\bibliographystyle{IEEEtran}
\balance 
\bibliography{biblio_dataset_Explanation}
\end{document}